\documentclass[10pt]{article}
\usepackage[utf8]{inputenc}
\usepackage{authblk}
\usepackage{setspace}
\usepackage[margin=2cm]{geometry}
\usepackage{graphicx}
\usepackage{subcaption}
\usepackage{amsmath}
\usepackage{lineno}
\usepackage{booktabs}
\usepackage{hyperref}
\usepackage{makecell}
\usepackage[labelfont=bf]{caption}
\usepackage{threeparttable}
\title{Deep Learning with Information Fusion and Model Interpretation for Health Monitoring of Fetus based on Long-term Prenatal Electronic Fetal Heart Rate Monitoring Data}

\author[1,2,$\dag$]{Zenghui Lin}
\author[3,4,5,$\dag$]{Xintong Liu}
\author[4,5]{Nan Wang}
\author[6]{Ruichen Li}
\author[4,5]{Qingao Liu}
\author[2]{Jingying Ma}
\author[6,*]{Liwei Wang}
\author[4,5,*]{Yan Wang}
\author[2,7,*]{Shenda Hong}

\affil[1]{School of Public Health, Peking University, Beijing, China}
\affil[2]{National Institute of Health Data Science, Peking University, Beijing, China}
\affil[3]{Peking University First Hospital, Beijing, China}
\affil[4]{Department of Obstetrics and Gynecology, Peking University Third Hospital, Beijing, China}
\affil[5]{National Clinical Research Center for Obstetrics and Gynecology (Peking University Third Hospital), Beijing, China}
\affil[6]{School of Intelligence Science and Technology, Peking University, Beijing, China}
\affil[7]{Institute of Medical Technology, Health Science Center of Peking University, Beijing, China}
\affil[*]{Corresponding authors. Email: hongshenda@pku.edu.cn (Shenda Hong), wjgqhn@263.net (Yan Wang), wanglw@pku.edu.cn (Liwei Wang)}
\affil[$\dag$]{These authors contributed equally to this work.}

\date{}

\newcommand{\gterm}[2]{\textbf{#1} & #2 \\}

\begin{document}

\maketitle

\begin{abstract}
Long-term fetal heart rate (FHR) monitoring during the antepartum period, increasingly popularized by electronic FHR monitoring, represents a growing approach in FHR monitoring. This kind of continuous monitoring, in contrast to the short-term one, collects an extended period of fetal heart data. This offers a more comprehensive understanding of fetus's conditions. However, the interpretation of long-term antenatal fetal heart monitoring is still in its early stages, lacking corresponding clinical standards. Furthermore, the substantial amount of data generated by continuous monitoring imposes a significant burden on clinical work when analyzed manually. 
To address above challenges, this study develops an automatic analysis system named LARA (Long-term Antepartum Risk Analysis system) for continuous FHR monitoring, combining deep learning and information fusion methods. LARA's core is a well-established convolutional neural network (CNN) model. It processes long-term FHR data as input and generates a Risk Distribution Map (RDM) and Risk Index (RI) as the analysis results.
We evaluate LARA on inner test dataset, the performance metrics are as follows: AUC 0.872, accuracy 0.816, specificity 0.811, sensitivity 0.806, precision 0.271, and F1 score 0.415. In our study, we observe that long-term FHR monitoring data with higher RI is more likely to result in adverse outcomes (p=0.0021).
In conclusion, this study introduces LARA, the first automated analysis system for long-term FHR monitoring, initiating the further explorations into its clinical value in the future.

\end{abstract}

\section{Introduction}
During pregnancy, hypoxia poses a significant risk to the fetus, causing irreversible damage to the neural system, fetal growth restriction (FGR), and even fetal death in severe cases \cite{chandraharan2007prevention}. Timely intervention during the prenatal period can improve fetal growth conditions in the uterus and prevent damage caused by hypoxia.
Identification is essential for intervention. When the fetus experiences hypoxia in the uterus, changes in FHR may occur, serving as a method to identify fetal hypoxia. Therefore, monitoring FHR during pregnancy plays a pivotal role in safeguarding fetal health.

Currently, intermittent FHR monitoring is the primary clinical approach. This conventional short-term monitoring typically lasts less than 40 minutes, limiting the duration of monitoring \cite{american2021antepartum}. However, the timing of fetal hypoxia is often unpredictable, rendering short-term FHR monitoring insufficient for capturing hypoxic events.
The advent of electronic fetal heart rate (eFHR) monitoring allows for high-quality continuous monitoring day and night. This method has proven advantageous for prolonged monitoring \cite{kapaya2019portable}. In contrast to the intermittent one, long-term FHR monitoring extends the duration, improving the ability to capture hypoxic events. This results in more comprehensive and reliable monitoring, assisting obstetricians in better evaluating the overall fetal status \cite{pieri2001compact}. Previous studies have verified the feasibility and acceptability of long-term monitoring for pregnant women \cite{crawford2018mixed}. 
However, interpreting long-term monitoring manually poses a significant burden on obstetricians due to the large amount of data involved.

Machine learning (ML) methods are extensively employed to aid in the interpretation of short-term FHR \cite{nunes2016computer,spilka2012using}. Machine learning algorithms for FHR monitoring can be broadly categorized into two types: traditional ML methods and deep neural networks (DNNs). Traditional ML methods have been proven effective in previous studies. Czabanski et al. design an expert system to predict neonatal acidemia using a two-stage analysis based on weighted fuzzy scoring and least square support vector machine (LS-SVM). The performance achieves an accuracy of 92.0$\%$ and a quality index of 88.0$\%$ \cite{czabanski2012computerized}. Spilka et al. advocate sparse support vector machine (sparse SVM) classification, enabling the selection of a small number of relevant features for efficient fetal acidosis detection. The classification achieves a sensitivity of 73$\%$ and specificity of 75$\%$ \cite{spilka2016sparse}. Chen et al. introduce a Deep Forest (DF) algorithm, integrating Random Forest (RF), Weighted Random Forest (WRF), Completely Random Forest (CRF), and Gradient Boosting Decision Tree (GBDT). The approach attains an area under the curve (AUC) of 99$\%$, with an F1-score and accuracy of 92.01$\%$ and 92.64$\%$, respectively \cite{chen2021DF}. Although some researchers have attempted to optimize the feature extraction process for improved accuracy \cite{liu2023baseline}, conventional ML methods often heavily depend on extracting complex features, potentially overlooking some deep features hidden in the raw data.
In recent years, with advancements in artificial intelligence in the medical field, researchers have begun applying DNNs to the automatic analysis of eFHR, proposing various DNN models \cite{comert2019fetal,liu2021attention,zhao2019deepfhr,ogasawara2021deep,xiao2022deep,zhong2022ctgnet,li2018automatic,zhou2023improvement}. Unlike traditional ML methods, DNNs operate as end-to-end models, taking raw data directly as input without prior feature extraction. This allows them to fully extract deep features hidden in the data. For example, Zhao et al. introduce an 8-layer deep CNN framework for automatic prediction of fetal acidemia. The model achieves an AUC of 97.82$\%$, with accuracy, sensitivity, specificity, and quality index values of 98.34$\%$, 98.22$\%$, 94.87$\%$, and 96.53$\%$, respectively \cite{zhao2019deepfhr}. Ogasawara et al. introduce a constructed deep neural network model (CTG-net) utilizing a long short-term memory (LSTM) structure to classify the abnormal and normal groups from cardiotocography data. The model achieves an AUC of 73$\%$ \cite{ogasawara2021deep}. 

Nevertheless, there are three main limitations with existing DNN models for eFHR. 
\begin{itemize}
\item First, previous studies mainly focus on the classification of short-term eFHR monitoring. Long-term eFHR monitoring serves as a brand-new prenatal care monitoring type, better reflecting the ongoing condition of intrauterine fetus. However, the extended duration of long-term monitoring poses a great challenge for analysis using existing models designed for short-term monitoring.
\item Second, most existing models lack interpretability analysis, which makes it difficult for obstetricians to use in the clinics. Different from traditional machine learning models, DNN construction is like a ``black box'', inexplicable for humans due to large amount of parameters and calculation \cite{XAI}. Therefore, enough interpretability analysis is crucial for controlling the bias and improving reliability of the model. 
\item Third, previous studies have scarcely explored the clinical value of long-term FHR monitoring. Although long-term FHR monitoring is anticipated to offer a more effective clinical approach, challenges in collecting long-term FHR data and the absence of convenient analysis methods make researching its clinical value challenging.
\end{itemize}

In this study, we gather a substantial long-term FHR monitoring database from Peking University Third Hospital. Building upon this, we introduce a Long-term Antepartum Risk Analysis System (LARA) that utilizes a CNN and an information fusion method for long-term FHR monitoring. We apply LARA to the collected data and explore the correlation between LARA's output and clinical outcomes. Our findings indicate that the RI generated by LARA can serve as a marker for identifying fetuses that are Small for Gestational Age (SGA). Furthermore, we attempt to delve into the deep features of LARA using various methods, such as Gradient-weighted Class Activation Mapping (Grad-CAM) and the Uniform Manifold Approximation and Projection (UMAP) algorithm. These approaches provide a better understanding of the outcomes produced by LARA.

The main contributions of our study are as follows: 
\begin{itemize}
\item We develop an automatic analysis system, LARA, for long-term FHR monitoring, combining the CNN model with information fusion methods. Three different operators (Basic, R-S, C-W) are developed for the information fusion part to comprehensively reflect the fetus's condition using LARA. To the best of our knowledge, this is the first study to establish a system for the interpretation of long-term FHR monitoring. 
\item We establish our model with sufficient interpretability, analyzing it from three perspectives: output analysis, attention visualization, and deep features analysis. All results suggest that our model has learned the relevant and reasonable features when analyzing FHR.
\item We preliminarily explore the clinical value of long-term FHR monitoring in our collected database using LARA. It is observed that pregnant women with high RI are more likely to experience adverse fetal outcomes compared to those with low RI.
\end{itemize}

\section{Materials and Methods}

\subsection{Study Population and Follow-up}
This dataset was retrospectively collected from April 2014 to December 2018 in Peking University Third Hospital, containing 114 sequences of long-term monitoring from 86 singleton deliveries. The ones with evidence of fetal malformation during pregnancy were excluded after screening. Monitoring was employed from 28 to $41^{+6}$ weeks of gestational age. Each pregnant women received at least one continuous monitoring before laboring. The fetal heart rate was presented in the form of points 4 times per second. Those who lost more than 50$\%$ of the FHR signals during monitoring were excluded. The raw fetal heart data points are exported through the device ``fetal/maternal heart rate recorder (AN24, Monica Healthcare, UK)'' and the corresponding software. The ethical approval number of this project is M2022195.

For pregnant/postpartum women, we collected their basic information encompassing age, gestational weeks for monitoring, gestational weeks at delivery, as well as mode of delivery. Maternal complications or comorbidities were recorded, such as hypertensive disorders of pregnancy, fetal growth restriction, and diabetes complicating pregnancy or gestational diabetes. Additionally, we documented information on the newborns after laboring, including birth weight, body length, gender, and Apgar scores \cite{apgar1953proposal}. Some newborns also underwent cranial ultrasonography. Furthermore, we assessed infants aged 0-6 months at the neonatal care center using the Peabody Developmental Motor Scale-2 (PDMS-2) \cite{folio1983peabody}. This was administered by healthcare professionals, in order to evaluate whether their motor development was within the normal range.

\subsection{Labeling Process}

The labeling process is conducted by two experts with more than 20 years of obstetrical experience. After assessment of each 20-minute segment according to non-stress test (NST) criteria, they apply a refined multi-class classification scoring system, which is more close to clinical practice. The scoring rubric could be seen in Table 1. Segments are divided in normal ones with scores greater than 3, and abnormal with score equal or less than 3.

\begin{table}[htbp]
\centering
\begin{threeparttable}
\caption{\textbf{Scoring Criteria}}
\begin{tabular}{ll}
\toprule
\textbf{Classification} & \\
\midrule
5 &  \makecell[tl]{ normal EFM\tnote{*}:\\  \hspace{0.5cm} \textbullet\ baseline 110-160 bpm;\\    \hspace{0.5cm} \textbullet\ variability 6-25 bpm;\\
   \hspace{0.5cm} \textbullet\ acceleration \textgreater 2 accelerations with acme of \textgreater 15 bpmlasting 15 sec.;\\  \hspace{0.5cm}  \textbullet\ deceleration none}\\
4 & atypical EFM\\
3 & atypical EFM\\
2 & abnormal EFM\\
1 & abnormal EFM\\
\midrule
\textbf{Deduction}&\\
\midrule
1 & baseline \textless 110 bpm or \textgreater 160 bpm \\
1 & variability $\leq$ 5 for 10 min.\\
2 & variability $\leq$ 5 for 20 min.\\
1 & variability \textgreater 25 bpm for 20 min.\\
1 & acceleration \textless 2 accelerations or acme of \textless 15 bpm, lasting \textless 15 sec.\\
1 & variable decelerations 30-60 sec. duration (each 1 score, could be accumulated)\\
2 & deceleration lasting \textgreater 3 min. or Late deceleration (each 2 score, could be accumulated)\\
\bottomrule
\end{tabular}
\begin{tablenotes}
\item[*] EFM: Electronic Fetal Monitoring
\end{tablenotes}
\end{threeparttable}
\end{table}

\begin{figure}[h]
    \centering
    \includegraphics[width=0.9\textwidth]{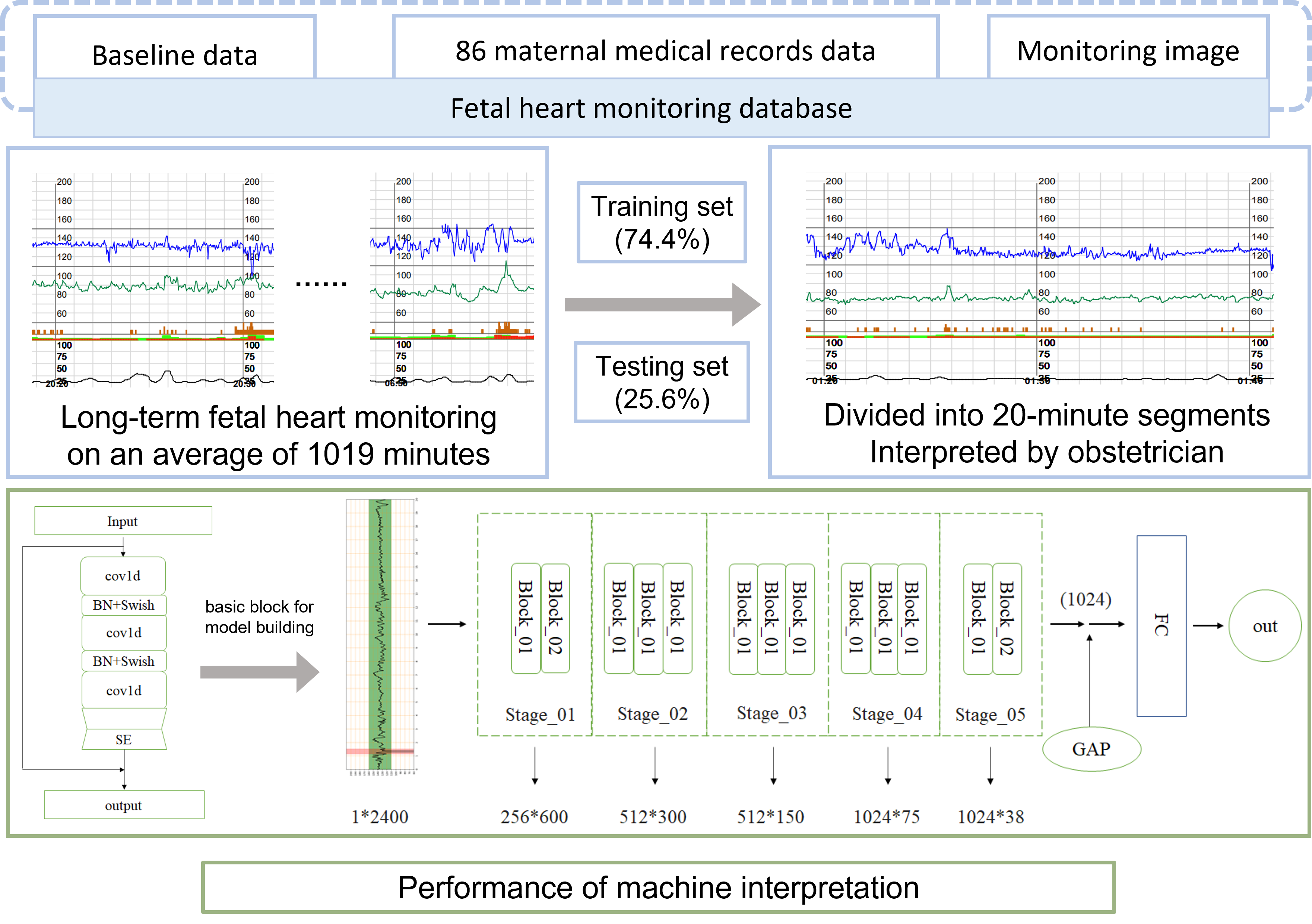}
    \caption{\textbf{Construction of prediction model}\\
   This figure illustrates the procedural steps involved in the construction of our prediction model. cov1d: Convolution layer designed for 1-Dimensional input. BN: Batch Normalization layer. Swish: An activation function. SE: Squeeze-and-Excitation layer. GAP: Global Average Pooling layer. FC: Fully Connected layer.}
    \label{Figure 1}
\end{figure}.

\subsection{Development of LARA}
To analyze the long-term monitoring data, we establish LARA based on deep CNN and information fusion methods. LARA mainly consists of three parts: Preprocessing, CNN model and results infusion. The system's core is a well-performing CNN model trained on long-term monitoring data. LARA takes long-term data as input and produces a RDM by analyzing the data through its three components. Additionally, LARA calculates the RI for the entire monitoring based on the RDM, offering a comprehensive assessment of long-term monitoring.

\subsubsection{Data Preprocessing}
Preprocessing is a crucial step in most biomedical signal processing. In LARA, preprocessing serves two primary purposes: controlling the quality of input data and preparing data for model training.

During clinical data collection, several factors can potentially affect data quality, leading to interference or even missing data. Examples include the movement of the mother and fetus, transducer displacement, and external clinical environmental factors. Previous studies commonly opt for different interpolation methods to fill the gaps caused by missing data or rewrite segments with low signal quality \cite{zhao2019deepfhr,zhong2022ctgnet}. However, to preserve the integrity of the original eFHR data, we exclusively fill the missing data points with zeros without employing additional interpolation methods. Long gaps (\textgreater10 minutes) are promptly removed at the outset.

Employing the stratified random partitioning method, the 114 long-term fetal heart monitoring records included in the study are split into a training set (91 monitoring, 79.8$\%$) and a testing set (23 monitoring, 20.2$\%$), maintaining a ratio of approximately 8:2. Because of the distinctive nature of medical data, DNNs employed in the medical field often encounter challenges related to imbalanced datasets. To address this issue, we resample the training data into 10-minute fragments and implement different steps for data with different labels. Specifically, for data labeled as normal, the step is set to 10 minutes, while for data labeled as abnormal and atypical, the step is reduced to 1 minute. Through resampling, we manage to alleviate the impact of dataset imbalance on model training (6.51$\%$, 208:2988). Following resampling, the proportion of abnormal and atypical data in the training dataset increases to 27.69$\%$ (abnormal: normal 2288:5976). However, in the testing dataset (7.35$\%$, 124:1564), we opt not to use resampling.

\subsubsection{Deep Learning Architecture}
Our model is based on a deep CNN that had been verified in many one-dimension medical time-series data tasks \cite{hong2020holmes}. While CNNs are traditionally designed for 2-D input data, 1-D CNNs have demonstrated excellent performance in medical time-series tasks in recent years \cite{1DCNN}. The calculation process in 1-D CNN mirrors that of conventional 2-D CNNs. In 1-D CNN, the input layer employs a specialized 1-D CNN kernel, maintaining a fixed input channel of one.
In summary, our model comprises five main stages and a final output layer. Each stage is constructed using basic convolution blocks, with the number of blocks in the stages being 2, 3, 3, 3, and 2, respectively (\autoref{Figure 1}). 

\textbf{Basic Block}. The structure of the basic block is shown in Figure 1. The basic block primarily consists of three convolution layers, with a Batch Normalization (BN) layer and a Swish activation layer utilized between two adjacent convolution layers. Our basic block adopts a residual architecture and integrates a squared-and-excitation (SE) attention mechanism. Deep CNNs frequently encounter the problem of gradient vanishing, impeding training efficiency. The incorporation of residual blocks, constructed with residual connections, has proven effective in addressing this issue and facilitating the training of deep CNN networks. It is worth noting that our residual connections feature SE layers. Squeeze-and-Excitation (SE) is an attention mechanism introduced by Jie Hu et al. in 2018 to enhance the performance of deep neural networks \cite{hu2018squeeze}. The SE module dynamically adjusts the weights of each channel, enabling the network to prioritize crucial features. This mechanism enhances both the expressive capability and overall performance of the network.

\textbf{Stage}. Building upon the basic blocks, we construct stages. In each stage, we have the flexibility to choose the number of basic blocks, providing the capability to adjust the depth and structure of the model. Moreover, in designing our stages, each basic block within a stage almost shares hyperparameters. This implies that we can swiftly modify the overall network structure and depth by directly adjusting at the stage level, facilitating rapid iteration and optimization of the network. 
At the end of the model, we use a Global Average Pooling (GAP) layer followed by a Fully Connected (FC) layer as the output layer. The GAP layer is commonly employed in CNNs as the output layer to enhance the model's performance. Specifically, for an input feature map, the GAP Layer computes the average value for each channel across the entire feature map and produces a single value for each channel. Following the GAP layer, the model obtains a deep feature vector of length 1024, and the FC layer utilizes this vector to make the final prediction.

\subsubsection{Information Fusion for Long-term Analysis}
While CNNs have demonstrated power in various medical fields, they typically require fixed-length inputs due to their specific structure. However, in clinical settings, collected eFHR data often varies in monitoring time. In LARA, we introduce a set of results infusion methods to address this challenge.

Results infusion is inspired by the concept of information infusion. Specifically, results infusion is implemented as follows: For long-term monitoring data, the system first divides it into 1-minute units. Subsequently, the CNN model scans this long-term data with a window size of 10 minutes in steps of 1 minute, recording the model prediction for each scanned minute-unit at different windows. Consequently, each minute-unit, except those at both ends, will be scanned by 10 windows, resulting in 10 predicted values for each minute-unit from different windows. To calculate the Risk Index of the minute-unit (mRI), the system combines the 10 predicted values of the same minute-unit using aggregating operators.
\begin{figure}
    \centering
    \includegraphics[width=1\linewidth]{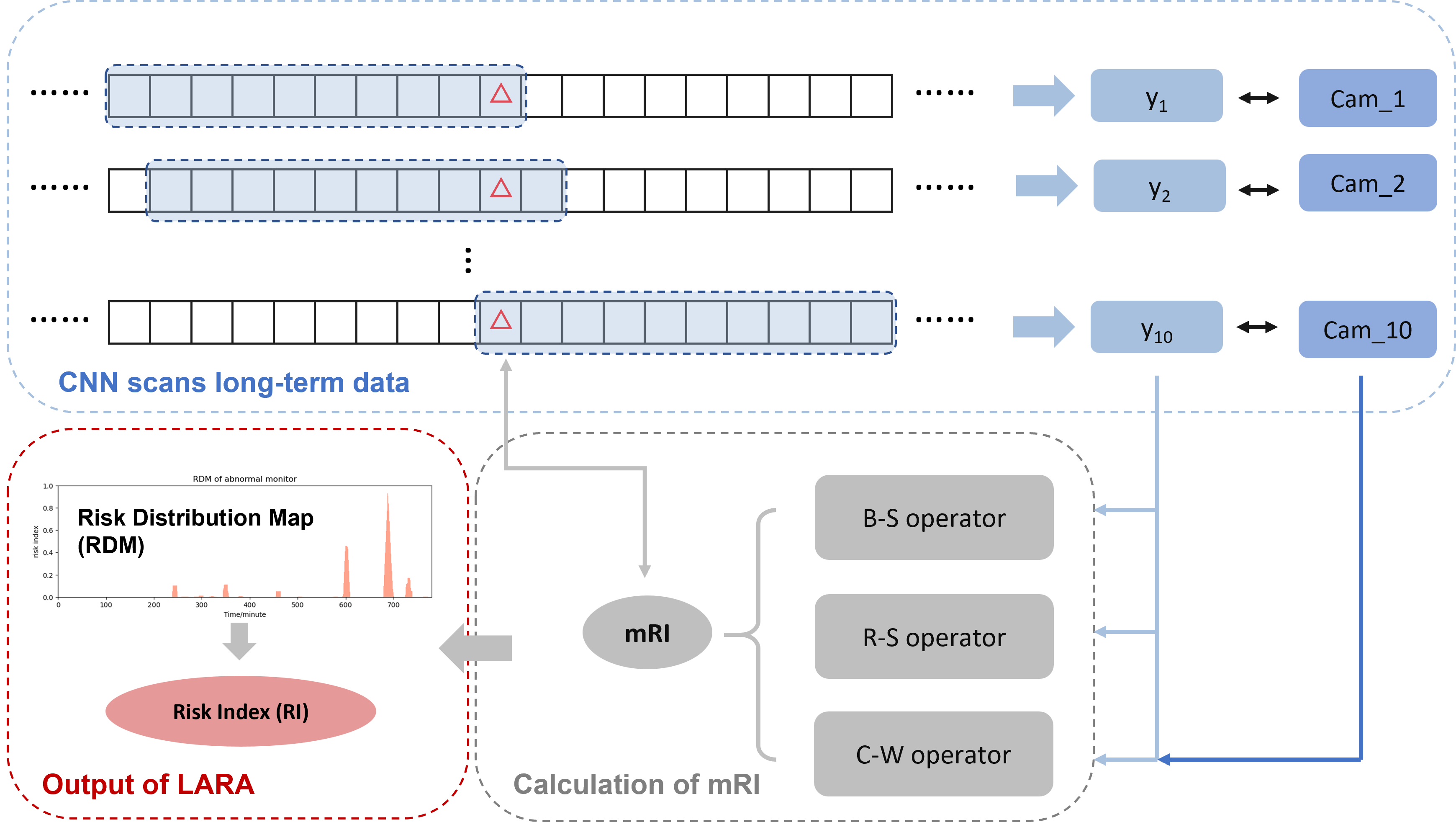}
    \caption{\textbf{Information Fusion process}\\ This figure delineates the stages involved in the process of information fusion. Cam$\_$i is calculated using the Grad-CAM algorithm. CNN: Convolutional Neural Network. mRI: risk index for minute-unit.}
    \label{Figure 2}
\end{figure}

In LARA, we introduce three different aggregating operators: Basic operator, Risk-Sensitive (R-S) operator, and Cam-Weighted (C-W) operator.

The Basic operator is a straightforward aggregating operator. Without incorporating any weights, the Basic operator directly averages the 10 predicted values recorded for each minute-unit to obtain its mRI.
$$Basic(x_1,x_2,x_3\cdots x_i)=\frac{\sum x_i}{i}$$

The R-S operator is devised with the aim of minimizing false negatives that might result in missed diagnoses. This operator is derived from the max function. However, unlike the max function, the R-S operator not only emphasizes higher values but also takes into account the contribution of each value being aggregated. Consequently, the R-S operator can fully leverage the data dimensions of all values, achieving improved risk-sensitive aggregation results.
$$R-S(x_1,x_2,x_3\cdots x_i)=\frac{\sum T_ix_i}{\sum T_i}$$
where $T_i=\text{e} ^ {(x_i- \Bar{x})}$ and $\Bar{x}=\frac{\sum x_i}{i}$

The C-W operator integrates the model's attention and results based on the Grad-CAM algorithm. Specifically, during the scanning of long-term data, we not only calculate the predicted value but also use the Grad-CAM algorithm to compute the attention distribution of the model. Since the input data has a particle size of 4Hz, the calculated cam value is also of the same particle size. Subsequently, the cam values of each minute-unit in a specific window are averaged. Finally, the predicted values are weighted with the obtained averaged cam values to calculate the mRI.
$$C-W(x_1,x_2,x_3\cdots x_i)=\frac{\sum x_iCam_i}{\sum Cam_i}$$
where $Cam_i$ is the average of the cam values of that specific minute-unit (\autoref{Figure 2}).

\vspace{\baselineskip}
\textbf{RDM}. Plotting the mRI, LARA generates the RDM with the X-axis representing time in minutes and the Y-axis representing mRI. The RDM essentially illustrates the mRI over time, depicting the distribution and changes of the mRI throughout the entire long-term monitoring process. The RDM serves as a comprehensive reflection of LARA's analysis results on long-term FHR data.

\textbf{RI}. From the RDM, LARA further calculates the RI for the entire long-term FHR, representing the area under the RDM. The formula for RI is as follows:
$$RI=\frac{\sum mRI}{T}$$
where $T$ represents the time of the FHR in minutes.

\subsubsection{Model Interpretation}
Neural network models are generally challenging for humans to interpret due to the vast number of parameters and calculations involved \cite{XAI}. Model interpretation has become a crucial aspect in understanding the inner workings of a model. In our study, we interpret the model from three perspectives.

\textbf{Output analysis}. The CNN model produces a predicted value between 0 and 1, reflecting the risk of a specific segment. We calculate the negative logarithm of the predicted values, resulting in predicted scores ranging from 0 to \(\infty\), for every segment. The predicted scores for all segments in the test dataset are calculated and compared with expert-labeled score values. Specifically, we examine the distribution in two classes (normal or abnormal) and five classes (score-value) of the predicted score. Additionally, we employ non-parametric tests to assess differences between different classes.
$$Predicted \_ score=-log_{10}(Predicted \_ value)$$

\textbf{Attention Visualization}. Grad-CAM is a widely used attention-visualization algorithm in 2-D CNN for Image Recognition Technology (IRT). We apply the same algorithm in our 1-D CNN model to visualize attention. During training, a CNN undergoes both forward and backward processes, with the latter involving calculating the gradients of every convolution kernel for optimization. When testing, backward processing is not performed as it is unnecessary. Similarly, Grad-CAM runs the backward process and restores the gradient during calculation. Grad-CAM then calculates the cam value for each data point using the stored gradient, representing the contribution of that point to the final prediction value \cite{selvaraju2017grad}. Visualization of the model's attention can be derived from these cam values.

\textbf{Deep Features analysis}. A deep CNN model undergoes multiple convolution layers to capture hidden features in the input data before producing the final prediction based on these features. Therefore, studying the deep features concealed within the CNN model allows us to gain a better understanding of what the model has learned during training and the reasoning behind its predictions. Our CNN model consists of numerous convolution layers, with each layer capturing distinct deep features and passing them to the next layer in the network. Among all the deep features, the layer just before the final output is considered the most connected to the output, as the final prediction relies on the deep features of that layer. For convenience, we extract the deep features of the last convolution layer before the output layer in our test dataset. In our CNN model, the last convolution layer consists of 1024 kernels, implying that the deep features of that layer form a vector of length 1024. Analyzing deep features of 1024 dimensions can be challenging, so we utilize the UMAP algorithm to extract the significant components of these deep features. UMAP is a dimension reduction algorithm employed to convert high-dimensional feature vectors in AI models into lower dimensions \cite{mcinnes2018umap}. In our study, we employ UMAP to convert the 1024-dimensional deep features into a 2-dimensional vector for further analysis. After plotting the 2-dimensional deep features, we select twelve segments representing four types of deep features: abnormal clustering, abnormal outliers, and their neighboring normal points.

\subsection{Implementation Details}

\subsubsection{Evaluations}
We employ various evaluation indicators to assess the effectiveness of the model, including AUC, accuracy, precision, sensitivity, specificity, and F1 score. The AUC represents the area under the ROC curve, which plots the performance of a binary classification model at different threshold settings. AUC reflects the overall classification ability of the model. The threshold is selected by maximizing the Youden Index. Based on the selected threshold, we calculate the accuracy $Accuracy  = \frac{TP+TN}{FP+FN+TP+TN}$, precision $Precision = \frac{TP}{TP+FP}$, sensitivity $Sensitivity(Recall) = \frac{TP}{TP+FN}$, specificity $Specificity = \frac{TN}{TN+FP}$, and F1-score $F1 = 2\times \frac{Precision\times Recall}{Precison+Recall}$ from the common binary confusion matrix. Where TP, FP, FN and TN represent true positive, false positive, false negative, and true negative respectively. In this work, the normal label (N) is considered negative, and the abnormal label (P) is positive.

\subsubsection{Statistics and Software} 
Baseline analyses and clinical validation are preformed using the IBM SPSS Statistics 26 software. Normal distribution data are presented in the form of mean ± standard deviation, while non-normal distribution data are described in terms of median (interquartile range). Among the former data, independent two sample t-test is applied. As for the latter one, we utilize Mann-Whitney U test, due to a small sample size of testing set (n=26). 2 × 2 contingency tables are tested by Pearson's chi-squared test. We also use Wilcoxon rank-sum test to compare between two groups. The CNN model is built using Python 3.10.10. All the plots are drawn using R 4.2.3.

\section{Results}

\subsection{Baseline materials}
\begin{figure}[htp]
    \centering
    \includegraphics[width=0.9\textwidth]{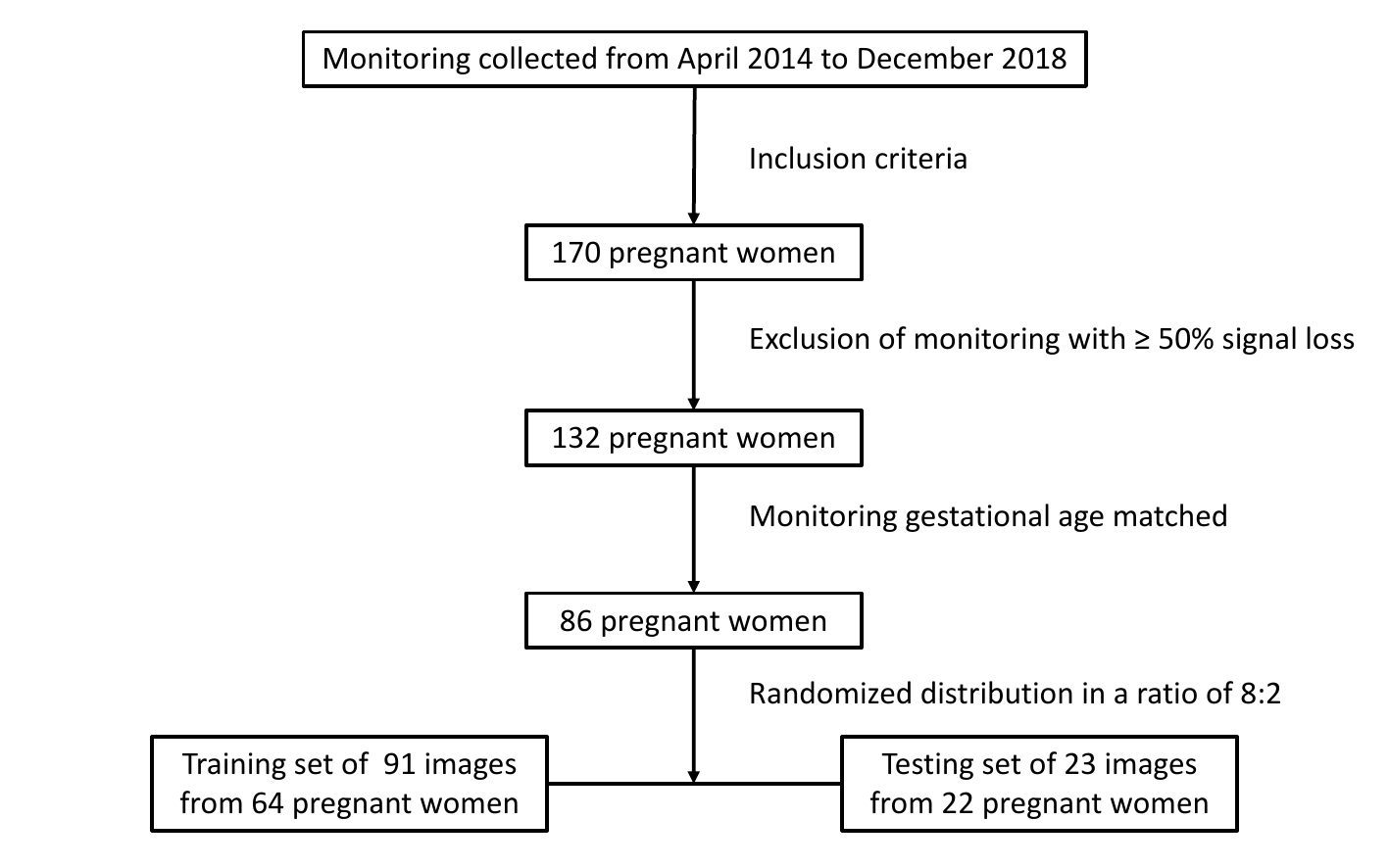}
    \caption{\textbf{Flow chart of normal and abnormal group recruitment}}
    \label{Figure 3}
\end{figure}

The dataset of 170 pregnant women was retrospectively attained from April 2014 to December 2018 in Peking University Third Hospital. After excluding monitoring with $\geq$ 50$\%$ signal loss, a total of 132 was included. For a matched baseline data, we excluded samples of \textgreater29 and \textless$37^{+6}$ monitoring gestational weeks. Each pregnant received at least one time of monitoring with different duration, with an average duration of 1019 minutes. Eventually, 114 long-term monitoring images from 86 singleton deliveries were included, without evidence of fetal malformation during pregnancy screening (\autoref{Figure 3}). 

114 fetal heart rate monitoring images included in the study are divided into a training set of 91 images from 64 singletons and a testing set of 23 images from 22 singletons (\autoref{baseline}). This indicates randomness of dataset partition. 

\begin{table}[htbp]
\centering
\begin{threeparttable}
\caption{\textbf{General characteristics of training and testing set}}
\begin{tabular}{cccc}
\toprule
\textbf{General characteristics} & \textbf{Training set (N = 64)} & \textbf{Testing set (N = 22)} & \textbf{p-value}\\
\midrule
Age(y)& 32.9$\pm$ 4.3& 32.3$\pm$4.5& 0.575\\
Gestational age of delivery (w)&  37.0(35.3,39.0)&  38.0(37.0,40.0) &  0.146\\
Birth weight(g)& 2603.3$\pm$704.0& 2828.6$\pm$777.0& 0.211\\
Vaginal delivery & 32(50.0) & 11(50.0)&\\
Cesarean & 32(50.0) & 11(50.0)&\\
Neonatal sex & n($\%$) & n($\%$) & 0.461\\
male & 32(50.0) & 13(59.1)&\\
female & 32(50.0) &9(40.9)&\\
Number of monitoring images & 91 &23&\\
Monitoring gestational age(w)&33.9(32.0,36.0)&34.4(33.0,36.0)&0.418\\
Monitoring time (min)&1023.4(886.7,1165.0)&1003.3(823.3,1125.0)&0.606\\
\bottomrule
\end{tabular}
\begin{tablenotes}
    \item Age and Birth weight are expressed as mean ± standard deviation, with an independent two-sample t-test employed for statistical analysis. Gestational age of delivery, Monitoring gestational age, and Monitoring time are represented using median (interquartile range), and statistical analysis is conducted using the Mann-Whitney U test. Neonatal sex is assessed using Pearson's chi-squared test.
\end{tablenotes}
    \label{baseline}
\end{threeparttable}
\end{table}

\subsection{Evaluation of model’s performance for short-term classification}
\begin{figure}
    \centering
    \includegraphics[width=0.9\linewidth]{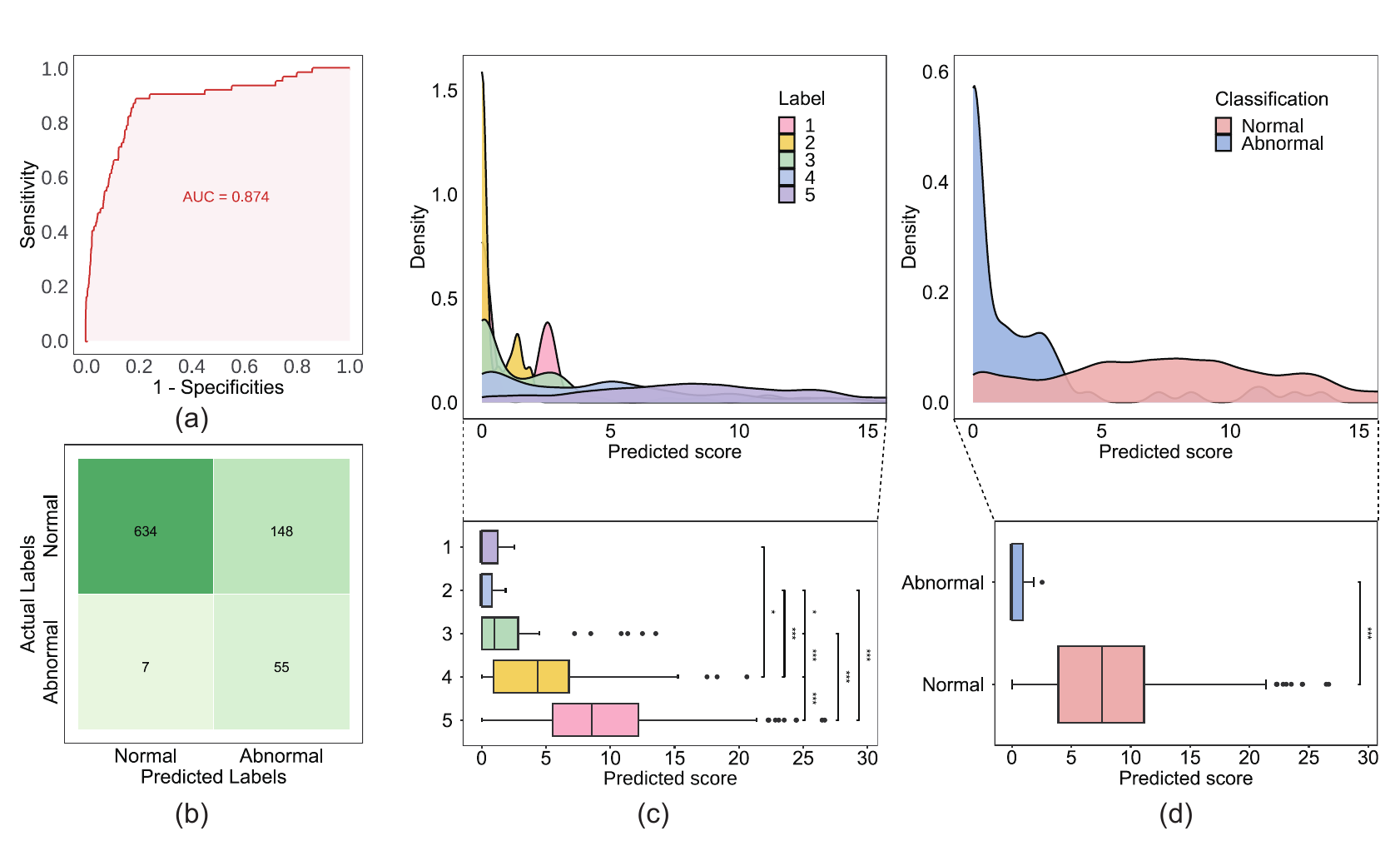}
\caption{\textbf{Model performance}\\ 
(a) Receiver operating characteristic (ROC) curve of the convolutional neural network (CNN) model for the test dataset. (b) Confusion matrix depicting the classification outcomes of the model, in conjunction with the threshold applied based on the Youden Index. (c) (d) Distribution of predicted scores among manually labeled five groups and two groups, illustrated through the density plot and boxplot, respectively. In (c), the five groups pertains to the categorization based on manually assigned scores according to the Scoring rubric outlined in Table 1. In (d), the classification criterion for normal is established as scores greater than 3, whereas abnormal is defined as scores equal to or less than 3.}
\label{Figure 4}
\end{figure}

As the results are interpreted in segments, we first analyze the performance in 20-minute segments. Utilizing the test dataset, we plot the ROC curve for segment classification and calculate the AUC (\autoref{Figure 4}a). Additionally, accuracy, specificity, precision, and F1 score are employed to evaluate the model's performance. Before resampling, the AUC, accuracy, specificity, sensitivity, precision, and F1 score of the model were 0.773, 0.823, 0.829, 0.710, 0.193, and 0.303, respectively. To mitigate the impact of an imbalanced training dataset, we resample the 20-minute segments and retrain our model. Ultimately, our model achieves improved performance. Following resampling, the AUC, accuracy, specificity, sensitivity, precision, and F1 score of our model are 0.872, 0.816, 0.811, 0.806, 0.271, and 0.415, respectively (\autoref{performance}). 
To enhance our understanding of the model output, we employ a confusion matrix to illustrate the distribution of model results (\autoref{Figure 4}b). Analyzing the predictions of the model on the test dataset, it is evident that errors primarily occur in misclassifying abnormal data as normal data.

\begin{table}[htbp]
\centering
\caption{\textbf{Model Performance}}
\begin{tabular}{ccccccc}
\toprule
\textbf{Model} & \textbf{AUC} & \textbf{Accuracy} & \textbf{Specificity} & \textbf{Sensitivity} & \textbf{Precision} & \textbf{F1}\\
\midrule
Pre-model& 0.773& 0.823& 0.829& 0.710& 0.193&0.303\\ 
Post-model&  0.872&  0.816&  0.811&  0.887&  0.271& 0.415\\
\bottomrule
\end{tabular}
    \label{performance}
\end{table}

\subsection{Clinical Outcome Association Analysis for long-term classification}
To investigate the clinical value and performance of long-term analysis with LARA, we apply the model to analyze long-term monitoring data from our dataset. This involves calculating RDM and Risk Index RI for each piece of data. Subsequently, we examine the relationship between RI and fetal outcomes.
\begin{figure}
    \centering
\includegraphics[width=0.9\textwidth]{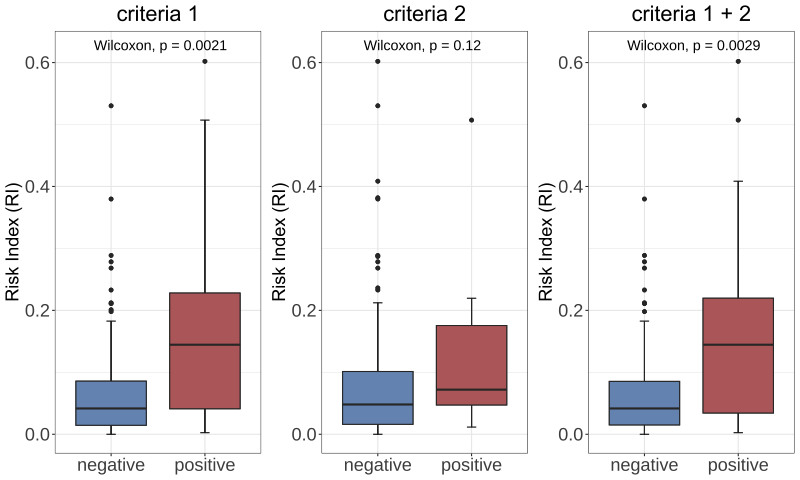}
    \caption{\textbf{RI validate for fetal outcomes}\\
    Distribution of Risk Index (RI) for long-term FHR across various fetal outcomes. Criteria 1 delineates a positive outcome as the manifestation of either maternal superimposed preeclampsia (SPE), the coexistence of chronic hypertension complicated with preeclampsia (PE), or the fetus falling below the 3rd percentile for Small for Gestational Age (SGA). For criteria 2, positive is defined as fetal brain disorder, including intraventricular hemorrhage, cerebral white matter softening, enhanced echogenicity in the cerebral white matter, and cerebral white matter loss. In the combined assessment of Criteria 1 and 2 (denoted as Criteria 1 + 2), a positive determination is established if either Criteria 1 or Criteria 2 is met.}
    \label{Figure:RIoutcomes}
\end{figure}

We track the birth outcomes of 86 pregnant women in our dataset and conduct a correlation test between the RI calculated by LARA and  two criteria of poor clinical indication. Criteria 1 is defined as the occurrence of either maternal superimposed preeclampsia (SPE), the presence of chronic hypertension complicated with preeclampsia (PE), or the fetus being less than the 3rd percentile for SGA. Criteria 2 is defined as fetal brain disorder, including intraventricular hemorrhage, cerebral white matter softening, enhanced echogenicity in the cerebral white matter, and cerebral white matter loss. P values are 0.0021, 0.12, respectively. Comparisons between those with both conditions and other populations reveal a significant difference in RI results (p=0.029). In (\autoref{Figure:RIoutcomes}), we depict the distribution of RI values calculated by LARA under the classifications of fetal brain injury and fetal growth restriction, respectively.
\subsection{Model Interpretation Results}

\subsubsection{Output Analysis results}

The distribution of outputs in two-class and five-class categories is illustrated in \autoref{Figure 4} c, d. For the two-class analysis, we test the difference of outputs between normal and abnormal groups and find that the outputs in the normal group are significantly higher than those in the abnormal group (P\textless0.001). In the five-class analysis, we observe that the mean output increases from score 1 to 5. Testing the differences of outputs among the five groups reveals a significant distinction between them (P\textless0.001). However, when testing between the score 1 and score 2 groups, we don't find a significant difference.

\subsubsection{Attention Visualization results} 

\begin{figure}[htp]
    \centering
    \begin{subfigure}{0.9\linewidth}
        \centering
        \includegraphics[width=0.9\linewidth]{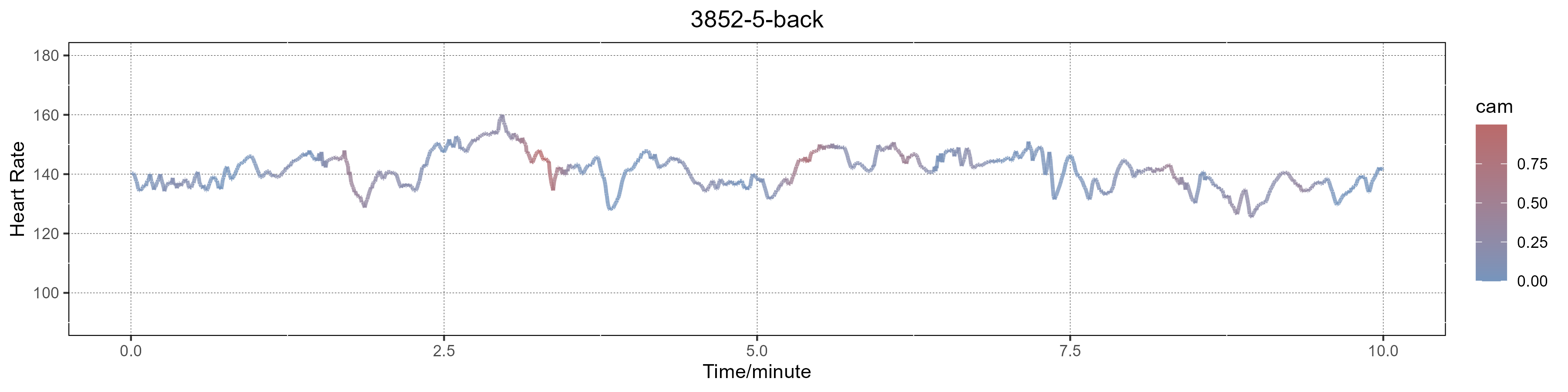}
        \caption{Recognition of features for normal monitoring}
    \end{subfigure}
    \qquad
    \begin{subfigure}{0.9\linewidth}
        \centering
        \includegraphics[width=0.9\linewidth]{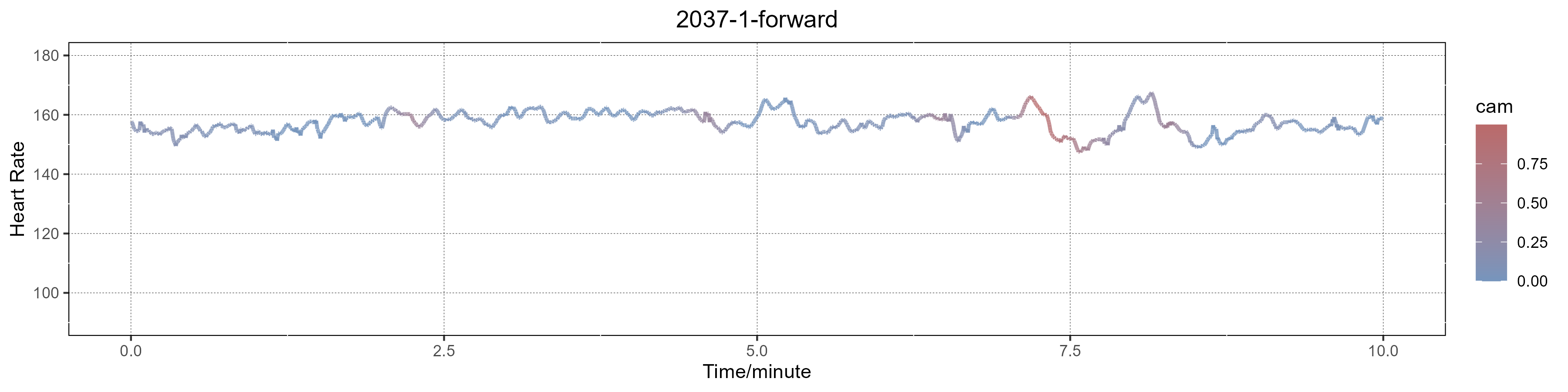}
        \caption{Recognition of features for abnormal monitoring}
    \end{subfigure}
    \caption{\textbf{Model attention visualization by Grad-CAM}\\
    This figure illustrates two instances of Grad-CAM visualization results, one corresponding to a segment of normal monitoring (a), and the other corresponding to abnormal monitoring (b). The cam delineates the attention of the model, with colorization specifically applied to the FHR. The intensity of red within a segment corresponds to the level of attention allocated by the model to that particular region. A deeper shade of red indicates a higher concentration of model attention in the corresponding FHR segment.}
    \label{Figure:gradcam}
\end{figure}

We choose two cases (a normal one and an abnormal one) from our dataset to visualize the attention of the model (\autoref{Figure:gradcam}). We color the 10-minute FHR data based on the cam value of each data point, allowing the color to reflect the attention of our model. Each data point's cam value ranges from 0 to 1. In Figure 6, the larger the cam value, the redder the segment. Redder segments attract more attention from our model when making predictions.
. 

\subsubsection{Deep Features analysis results}
\begin{figure}[htp]
    \centering
    \begin{subfigure}{0.9\linewidth}
        \centering
        \includegraphics[width=0.9\linewidth]{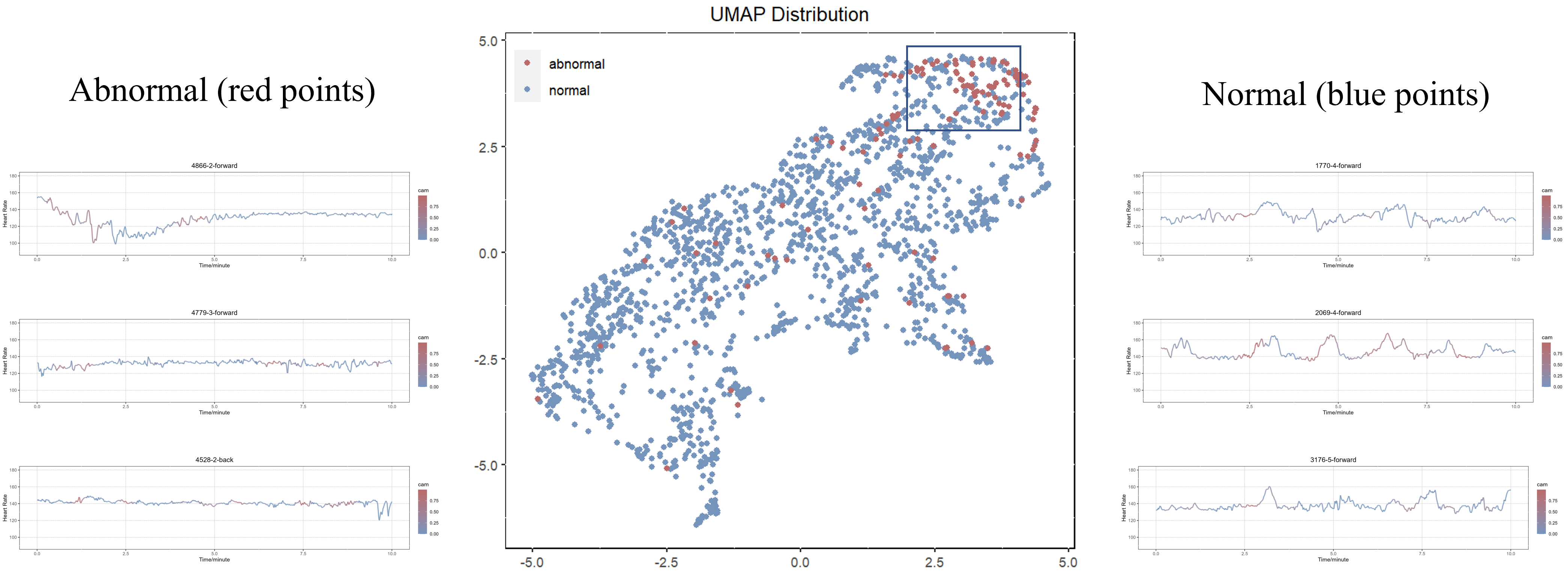}
        \caption{}
    \end{subfigure}
    \qquad
    \begin{subfigure}{0.9\linewidth}
        \centering
        \includegraphics[width=0.9\linewidth]{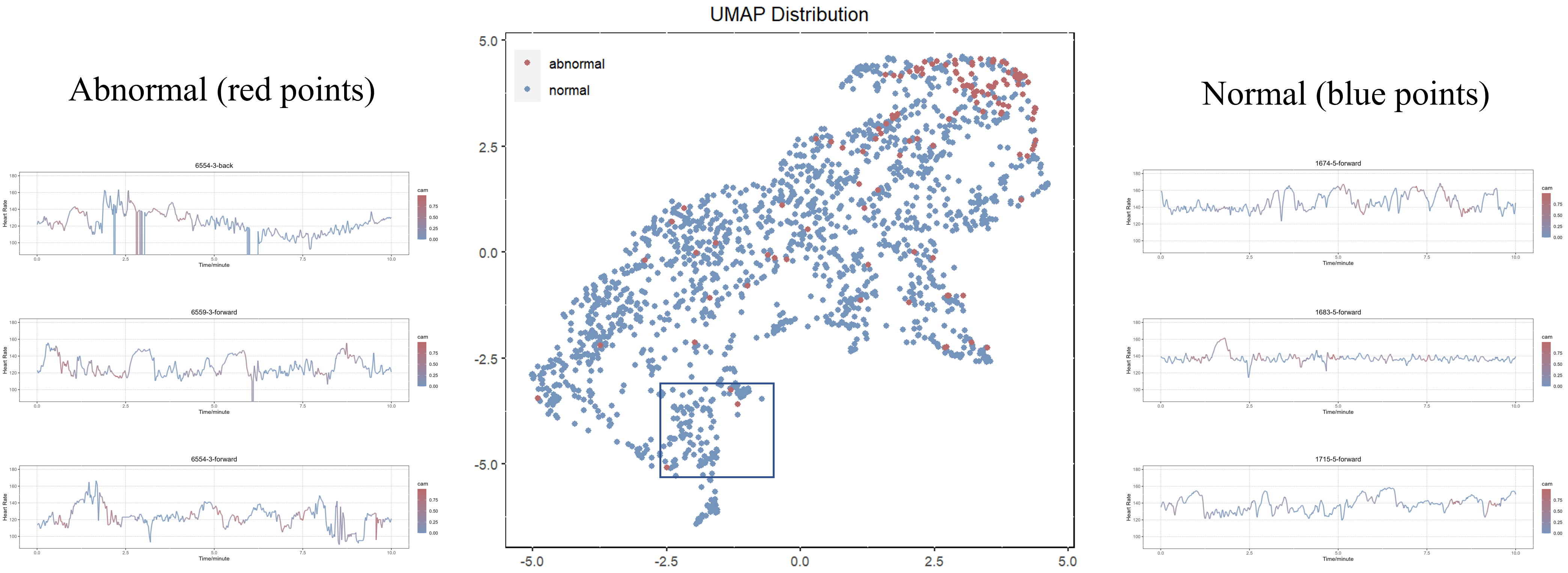}
        \caption{}
    \end{subfigure}
    \caption{\textbf{UMAP distribution and case study}\\
    This figure depicts the distribution of deep features within the test dataset following dimension reduction via the UMAP algorithm, along with twelve illustrative cases. In this representation, red points signify abnormal segments, while blue points denote normal segments. (a) Three randomly selected blue and red points are showcased from the upper-right corner of the UMAP, where a distinct clustering of red points is observed. (b) Three outlier red points, along with adjacent blue points, are chosen for case study in the bottom-left corner of the UMAP.}
    \label{Figure 7}
\end{figure}

In our UMAP distribution, red points (representing abnormal segments) display a clustering distribution in the upper right corner. However, many blue points (representing normal segments) are also distributed in that region, potentially contributing to false positive cases. We select three segments each of red and blue from the upper right corner to explore the reasons (\autoref{Figure 7}). 
. 
Additionally, there are some red outliers distributed in the bottom left corner, potentially contributing to false negative cases. We also select three segments of red points and three segments of nearby blue points (\autoref{Figure 7}). From the three red segments selected, high signal noise and low quality can be observed.

\subsection{Long-term analysis results}

In Figure 8, we extract two typical examples for presentation and comparison. It is observed that pregnant women with high RI are more likely to experience adverse fetal outcomes compared to those with low RI.
\begin{figure}[htp]
    \centering
    \includegraphics[width=0.9\textwidth]{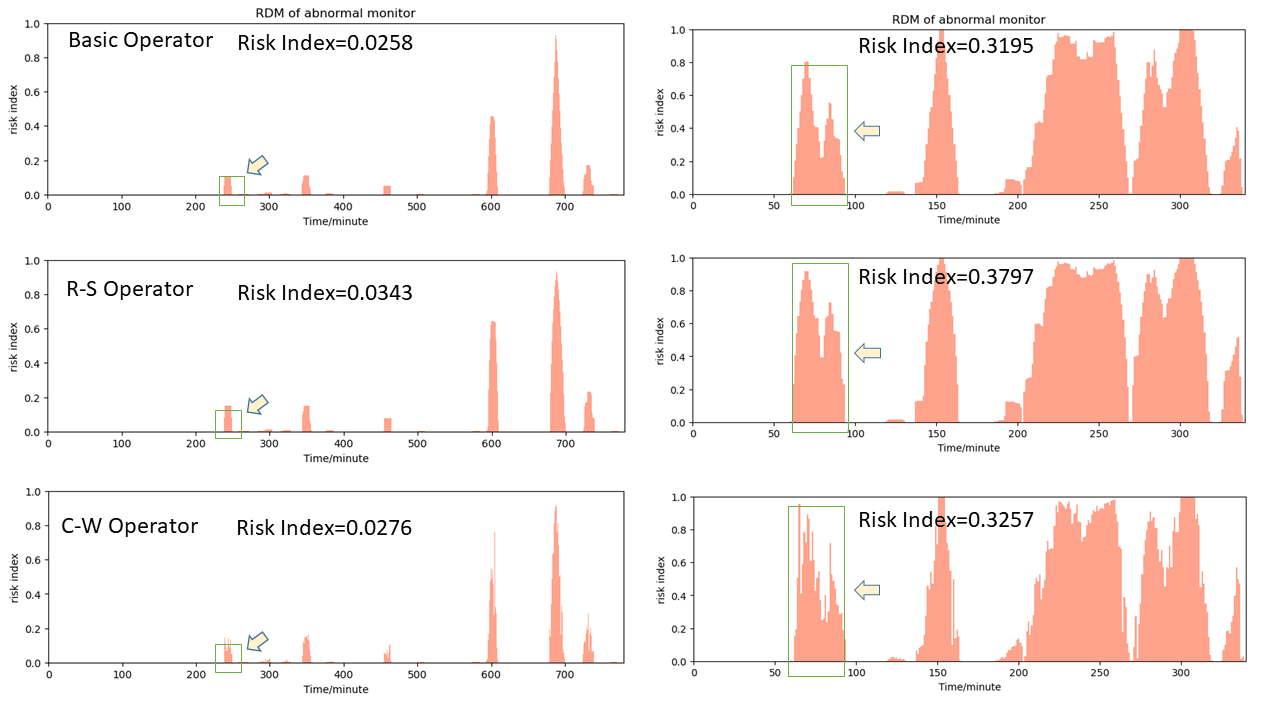}
    \caption{\textbf{Cases of RDM and RI}\\
    The Risk Distribution Map (RDM) and Risk Index (RI) using different operators for two typical examples selected from the test dataset. The highlighted region indicated by arrows is intended for the purpose of contrasting the disparities among the three operators.}
    \label{RDM}
\end{figure}

The left image is the RDM of a segment collected from a 25-year-old pregnant woman at 32 weeks of gestation. She had an uneventful pregnancy and delivered a healthy 3680g baby boy at full term (39 weeks) through a natural delivery. The baby had Apgar scores of 10 at 1 minute, 5 minutes, and 10 minutes after birth.

The segment of right image is from a 31-year-old pregnant woman at 31 weeks of gestation. A single male infant was born prematurely with a weight of only 1510g. The newborn had Apgar scores of 10 at 1 minute, 5 minutes, and 10 minutes after birth. The umbilical cord blood gas exhibited a pH of 7.36, and the infant was noted to be less than the 10$\%$ for Small for SGA. The newborn was transferred to the neonatal intensive care unit for observation and treatment for 10 days. During this period, the infant experienced complications, including episodes of apnea, patent ductus arteriosus, brain edema, jaundice, feeding intolerance, and inguinal hernia. Electroencephalogram (EEG) results showed normal brain activity. A head ultrasound revealed brain edema initially, but a follow-up examination indicated localized brain white matter injury on both sides after treatment.

\section{Discussion}
In this study, we develop an automatic analysis system (LARA) for long-term FHR monitoring using a CNN model. LARA not only facilitates the identification of intermittent segments but also enables long-term monitoring for risk identification by fusing information, achieving an AUC of 0.872. This provides an advanced reference model for clinicians. Additionally, leveraging LARA, we explore the clinical value of long-term FHR monitoring data. The interpretability of our model is demonstrated through the visualization of the system, addressing the subjective nature of FHR monitoring data interpretation. Moreover, we present a set of feasible long-term evaluation indicators.

The core of LARA is the well-trained CNN model, and the accuracy of this model directly influences the overall interpretation quality of LARA. Across various evaluation metrics for model performance, our model achieves an AUC of 0.872. Additionally, both Specificity and Sensitivity reaches 0.811 and 0.887, respectively, after threshold adjustment based on the Youden index. However, the model's performance is suboptimal in terms of Precision and F1-score, suggesting a higher proportion of false positives in its prediction results. This observation is corroborated by the confusion matrix. In the test dataset, there are only 62 positive (abnormal) data instances, but our model predicted 202 positives, with 148 of them being false positives. We attribute this discrepancy mainly to the low proportion of positive data in our test dataset (7.35$\%$), as we don't resample the test dataset as we do with the training dataset. It's worth noting that although we use binary labels during training, our model perform well when compared to 5-class labels, effectively distinguishing between classes 4, 5, and abnormal. This observation suggests that our model has learned the relevant features and can make accurate classifications of FHR data. However, our model shows less effective discrimination within the abnormal groups 1, 2, and 3. If more abnormal data can be collected in the future, it may enhance the model's ability to classify abnormal data more clearly.

From our UMAP result, it can be observed that most of the red dots representing abnormal data are clustered in the upper-right corner, indicating that the deep features learned by the model have good discriminative properties for abnormal data. In order to understand the reasons behind the model's errors, we randomly select three data points that are farther away from the cluster for observation (\autoref{Figure 7}). It can be seen that these outlier data points generally exhibit significant fluctuations, possibly due to noise introduced during data collection, such as electrode movement or maternal movement. Furthermore, from the UMAP distribution, it is evident that while abnormal data points are clustered, there is no clear boundary between nearby normal and abnormal data points, corroborating the previously mentioned issue of a high false-positive rate in the model. To understand the reasons behind the model's false positives, we also select three data points for observation. It can be observed that false positives often occur in normal data with more instances of fetal heart accelerations, and the model's attention is also focused on the acceleration segments. Conversely, in correctly identified normal data, not only accelerations are evident but also noticeable deceleration, which might be a contributing factor to the model's tendency to produce false positives.

In the Grad-CAM results, the model's attention is concentrated on the deceleration part of fetal heart monitoring (\autoref{Figure:gradcam}). We compare segments correctly identified by the model as normal and abnormal. It can be observed that, irrespective of whether the segment is normal or abnormal, the model's focus is on the acceleration and deceleration parts of these two data. Acceleration and deceleration are often crucial features analyzed by manual and traditional machine learning methods in the interpretation of FHR \cite{de2019automated}. In our results, there are more deceleration segments in normal data, exhibiting higher heart rate variability. In contrast, in abnormal data, the heart rate variability is lower, and the identifiable deceleration segments are relatively scarce.

Results fusion is the final and crucial component of LARA. We have designed three information fusion operators for LARA: the Basic Operator, R-S Operator, and C-W Operator. When considering the overall RI, the three operators perform with no significant difference, and the RI consistently performs as R-S > C-W > Basic. R-S, as a risk-sensitive operator, maintains a higher risk index. The C-W operator, which derives information integration results based on model attention weighting, being greater than the Basic operator, may suggest that the model tends to pay more attention to data with higher risk, indirectly indicating that the model's attention is correct. Looking into the details, the C-W operator appears to reflect finer-grained information. In Figure 8, the mRI for the Basic Operator and R-S Operator shows some continuity, meaning that the difference between nearby two-minute windows is relatively small. In contrast, the C-W operator performs a jumping mRI, which may suggest that the C-W operator can provide more detailed information on mRI.

Our research explore the clinical value of long-term FHR monitoring for the first time. In our study, it is found that the RI calculated by the LARA show significant differences between the group with SGA and the group without SGA (P=0.0021). This implies that pregnant women with higher RI are more likely to experience SGA. It suggests that the RI values calculated by the LARA may faithfully reflect the fetal growth status in the mother's body and may serve as a marker for screening SGA.
While FHR monitoring is originally a clinical method for screening fetal growth restriction, short-term monitoring often falls short in providing a comprehensive reflection of fetal growth due to its limited duration. Our study suggests that long-term monitoring data may achieve better monitoring results. The research also examine the correlation between fetal brain disorder and RI, but no significant correlation was found (P=0.12). This may be related to the process of RI value calculation or the inherent lack of correlation between the two variables. Exploration of the clinical value of long-term FHR monitoring data has just begun, and we hope that our work will inspire and assist future research in this field.

In recent years, numerous researchers have delved into the utilization of smart home monitoring devices to enhance maternal health during pregnancy \cite{appsurvey, laboronset}. A preceding study has also conducted tests on a self-guided fetal heartbeat monitor in comparison to cardiotocography \cite{monitoringdevice}. When considering the application of digital health interventions for pregnancy, the imperative lies in delivering tailored and personalized interventions \cite{personalization}. LARA, functioning as an automatic long-term FHR analysis system, facilitates extended periods of home monitoring and harbors the potential to provide personalized analyses of maternal health when integrated with home monitoring devices. This prospect delineates a promising avenue for future research endeavors dedicated to advancing maternal health during pregnancy.

This study has several limitations, mainly in the following aspects:
1) Due to the limited widespread adoption of long-term FHR monitoring, we don't validate the performance of LARA on a prospective dataset. Future research could benefit from validating the system on a larger and more diverse dataset to assess its generalizability.
2) Since our data is collected from clinical sources, even though we perform resampling, the bias in the training dataset can still influence the model's performance. Efforts to collect a more balanced and representative dataset would enhance the reliability of the developed system.
3) While this study has established a system for long-term FHR monitoring and established connections with neonatal outcomes, the clinical significance of long-term FHR monitoring interpretation remains somewhat limited. Further research should aim to explore and establish the clinical utility of long-term FHR monitoring in improving pregnancy outcomes.
4) Fetal development in the womb involves two distinct states, wakefulness and sleep, and these states can influence FHR monitoring results. This factor is not considered during the model development in this study. Future research could explore identifying different fetal wakefulness states from long-term FHR monitoring data, allowing for the analysis of FHR data based on these states. This approach may provide a more comprehensive interpretation of long-term FHR monitoring.
Addressing these limitations would contribute to the robustness and applicability of the automated long-term FHR monitoring system.

\section{Conclusion}

In this study, we develop an automated system for the analysis of long-term FHR monitoring data. This system can assist clinicians in rapidly analyzing long-term FHR monitoring data, thereby reducing the burden of manual analysis. Currently, long-term FHR monitoring is still in the exploratory stage, with limited widespread use and a lack of standardized interpretation criteria. Therefore, our proposed automated analysis system may help reduce the analysis costs associated with long-term FHR monitoring, thereby promoting its adoption. Since LARA is fully automated and requires no human intervention, it may be seamlessly integrated with portable monitoring devices. This enables home-based long-term FHR monitoring for pregnant individuals, further lowering the barriers to long-term FHR monitoring during pregnancy.

\section*{Code availability}

The detailed implementation code of LARA has been made publicly available on GitHub: https://github.com/goodboy-hub/LARA-FHR. The R code utilized for statistical analysis in this study is available upon request by contacting the corresponding author.

\section*{Acknowledgement}

This work was supported by the Beijing Natural Science Foundation (7232208), the Capital’s Funds for Health Improvement and Research (2022-2Z-40912), and the National Natural Science Foundation of China (62102008).

\section*{Table of Terminology}

\begin{table}[htbp]
\captionsetup{justification=raggedright,singlelinecheck=false}
\begin{flushleft}
\begin{tabular}{l p{0.6\textwidth}}
\toprule
\textbf{abbreviation} & \textbf{full title} \\
\midrule
\gterm{LARA}{Long-term Antepartum Risk Analysis System}
\gterm{RDM}{Risk Distribution Map}
\gterm{RI}{Risk Index}
\gterm{FHR}{fetal heart rate}
\gterm{eFHR}{Electronic fetal heart rate}
\gterm{AUC}{area under the receiver operating curve}
\gterm{CNN}{Convolutional Neural Networks}
\gterm{DNNs}{deep neural networks}
\gterm{mRI}{risk index for minute-unit}
\gterm{ROC}{receiver operating characteristic}
\gterm{F1}{F1 score}
\gterm{Grad-CAM}{gradient-weighted class activation map}
\gterm{UMAP}{Uniform Manifold
Approximation and Projection}
\gterm{CTG}{cardiotocography}
\gterm{FGR}{fetal growth restriction}
\gterm{SGA}{Small for Gestational Age}
\gterm{ML}{Machine learning}
\gterm{EFM}{Electronic Fetal Monitoring}
\gterm{GAP}{Global Average Pooling}
\gterm{FC}{Fully Connected}
\gterm{SE}{squared-and-excitation}
\gterm{BN}{Batch Normalization}
\gterm{AI}{Artificial Intelligence}

\bottomrule
\end{tabular}
\end{flushleft}
\end{table}

\bibliography{cite}
\bibliographystyle{unsrt}

\end{document}